\pdfoutput=1

\documentclass[11pt]{article}

\usepackage[]{ACL2023}

\usepackage{times}
\usepackage{latexsym}
\usepackage{graphicx}

\usepackage[T1]{fontenc}

\usepackage[utf8]{inputenc}

\usepackage{microtype}

\usepackage{inconsolata}

\newcommand{\lden}{[\![}
\newcommand{\rden}{]\!]}
\newcommand{\denotes}[1]{\lden #1\rden}
\title{Could the Road to Grounded, Neuro-symbolic AI \\ be Paved with Words-as-Classifiers?}

\author{Casey Kennington \\
Department of Computer Science \\
  Boise State University \\
  \texttt{caseykennington} \\
  \texttt{@boisestate.edu} \\\And
  David Schlangen \\
  Department of Lingusitics \\
  University of Potsdam \\
  \texttt{david.schlangen} \\
  \texttt{@uni-potsdam.de}}

\begin{document}
\maketitle
\begin{abstract}
Formal, Distributional, and Grounded theories of computational semantics each have their uses and their drawbacks. There has been a shift to ground models of language by adding visual knowledge, and there has been a call to enrich models of language with symbolic methods to gain the benefits from formal, distributional, and grounded theories. In this paper, we attempt to make the case that one potential path forward in unifying all three semantic fields is paved with the words-as-classifier model, a model of word-level grounded semantics that has been incorporated into formalisms and distributional language models in the literature, and it has been well-tested within interactive dialogue settings. We review that literature, motivate the words-as-classifiers model with an appeal to recent work in cognitive science, and describe a small experiment. Finally, we sketch a model of semantics unified through words-as-classifiers. 
\end{abstract}

\section{Introduction}


In his well-known \textit{Chinese Room} thought experiment, John Searle argued that the classical artificial intelligence (AI) methods of the day were not actually doing anything meaningful because they were working at the symbolic level, rather than having any connection with the real world \cite{Searle1980-wc}.\footnote{The thought experiment itself explains that a person who reads Chinese characters, looks up in a pre-defined book what the response character should be, and provides that response has no knowledge of the Chinese language and therefore isn't doing anything that requires any degree of intelligence.} This was extended a decade later by Stevan Harnad who, focusing on meanings of words, challenged the claim that meaning could be found in a formal, symbolic system because such a system is just an abstraction to the thing it is trying to model: ``meaningless symbols" connecting to other meaningless symbols \cite{Harnad1990-fr}. What was lacking, was a kind of \textit{grounding}, or direct reference to the real world---human cognition and therefore language is so tied to the physical world that computational models of semantics should be as well. 

More recently, a dramatic shift took place when classical, formal meaning representations were supplanted by distributional representations within the field of natural language processing and machine learning in general. The basic distributional theory is often attributed to JR Firth who observed that ``you shall know a word by the company it keeps" \cite{Firth1957-xy} and indeed words keep company with other words within speech and text, so something about how words show up in text must have something to do with their meaning. Meanwhile, textual datasets increased in size and extended across domains, giving the distributional hypothesis something empirical to work with. Simple co-occurrence methods were replaced by more principled models, such as word2vec \cite{Mikolov2015-it,Mikolov2013-su}, where the `meaning' of a word could be represented as computable vectors (i.e., points in n-dimensional space; see Figure~\ref{fig:distexample} for an example), amenable to the popular, neuro-inspired machine learning models and hardware, and trainable with text data. This was seen as a vast improvement over formal representations as word meanings were no longer intrinsic on the symbolic system; rather, the vectors and their distances from each other could be regarded as a kind of meaning representation that was computable. Indeed, in many circles, the wholesale use of the word \textit{embeddings} (i.e., the word vectors themselves derived from text data) became synonymous with \textit{meaning}. More principled, though computationally expensive models such as Attention neural network layers and Transformer architectures (which form the model that upon which most Large Language Models are built) enabled vector representations that go beyond the word level into the sentence and discourse level \cite{Vaswani2017-kv}, solidifying the supremacy of the distributional theory of semantics with empirical backing.

Surely the shift from formal, symbolic meaning representations to distributional theories that underlie computational semantics was for the better because it was so data-driven, it could scale, and evidence was mounting that it was effective at many, many tasks and benchmarks that require a substantial degree of language understanding \cite{Rogers2020-cm}. Some cried foul, however, pointing out that the data used to train the new models, while linguistic in nature, is not complete: by-and-large, transformer-based large language models (LLMs) are trained on \textit{text}, a data source that is easy to come by, but in the end fall victim to the same issues raised by Searle and Harnad, namely, that even the vectors are to a degree meaningless because they only refer to each other (a ``merry-go-round" of symbols pointing only to other symbols, ala Harnad), not to the physical world \cite{Lucking2019-ll,Bender2020-ck}.\footnote{Though, see \citet{Gubelmann2024-pu} for an argument that language models don't need symbol grounding.} 

Where does that leave us? There have been efforts to tie the visual world to the computable world of text-derived vectors, with clear degrees of success (see \citet{Fields2023-zv} for an overview of methods and early models), which is a step in the right direction. Another important trend is the argument that bringing formal semantics back into the neuro-inspired AI models (i.e., deep learning) is our only hope for proper intelligence that includes important abilities like inference and reasoning \cite{Valiant2008-lg}, even as the scale of datasets and size of models increases to absurd levels. \citet{Marcus2020-mg}, for example, argues that the path forward requires a ``hybrid, knowledge-driven, reasoning-based approach, centered around cognitive models." Thus a push for neuro-symbolic AI is underway, a version of AI where the power of formal systems is integrated into LMs trained on data with an increasing number of papers claiming they are addressing the neuro-symbolic challenge. We are thus still left with two problems: the symbol grounding problem, and the neuro-symbolic problem. 

In this paper, we make the assertion that the path to neuro-symbolic AI must also be amenable to symbol grounding---solving both problems is crucial, and should be done within a dialogue setting. That means that a proper model of computational meaning should be formal (i.e., symbolic), distributional, \textit{and} grounded in the physical world, learned through interactive dialogue. We limit our focus on word-level semantics.\footnote{We are agnostic to the machinery from formal or distributional models that take the semantics beyond the word level; both have merits and limitations.} At the heart of our hypothesis is the that the \emph{words-as-classifiers} (WAC) model can unify the three theories of computational semantics and lead to a neuro-symbolic model that grounds and benefits from distributional methods. 

We explain the WAC model in the following section and motivate it as the model of choice for unification, then explain how WAC has already been used to bridge grounded with formal, as well as grounded with distributional, including a small-scale experiment that shows how WAC is a promising model to be integrated more directly with LLMs. Finally, we sketch a model the unifies formal and distributional theories with WAC towards addressing the neuro-symbolic and symbol grounding problems.

\section{The Words-as-Classifiers Model: Word-level Grounded Semantics}

\citet{Larsson2013-po} explored how linguistic meaning is related to perception: the goal is for a computational model to know the perceptual meanings of words, for example the ``perceptual meaning of `panda' allows an agent to correctly classify pandas in her environment as pandas." Their work evaluated a simple left/right classifier that learned when a point on a screen was on the left or right side given coordinate information relative to the center. 


\citet{Kennington2015-ia} extended both the theoretical underpinnings and the domain of reference to real-world objects `perceived' by a camera. The authors explained that, for each word, a classifier is trained (e.g., logistic regression) on positive and negative examples of the word being used to refer to specific objects (hence, it needed a dataset with segmented objects and referring expressions made to those objects). For example, if the words \textit{the red box} is used to refer to object $object_1$, then the features of $object_1$ can be used as positive training samples for all three words in the referring expression. Negative examples can be drawn from objects that are not referred with those words. The result is a binary classifier for each word that, when features of a new object is given to it, yields a probability that represents the degree of belief that the object is a good `fit' for the word that classifier estimates. In this formulation, the authors claim that the object is the \textit{exension}, and the classifier itself (e.g., the coefficients along with the functional application) forms a computational \textit{intension}. This work was further extended in \citet{Schlangen2016-mz} which scaled the WAC model to the well-known ReferItGame benchmark of referring expressions to real objects in photographs of natural scenes \cite{Kazemzadeh2014-sw}, and introduced the \textit{words-as-classifiers} terminology. In the WAC model, word-level semantics is the primary focus, but also carries a provision for composition: words are composed into larger phrases by multiplying probabilities from each classifier in the referring expression (i.e., at the level of \textit{extension}) and finding the highest probable object. 

WAC has some nice properties that make it not only theoretically pleasing, but also practical and useful: (1) it is a model that can directly ground words into any modality, though most work grounds words into visual features (e.g., objects in photographs),\footnote{Though see \citet{Moro2018-in} that grounded WAC classifiers into simulated muscle activations, attesting the flexibility of WAC.} (2) only a small amount of training data is necessary to train classifiers \cite{Torres-Foncesca2022-ry}, and (3) its word-level operation means it can be used in fast, interactive (incremental) dialogue systems. One negative property is that words are trained somewhat independently of words around them, making principled composition into phrases and sentences a modeling challenge, which could potentially be overcome if WAC is unified with formal and distributional methods. 

\subsection{WAC Outside of Computational Semantics}

Here, we further identify different practical uses of WAC within neural network interpretation and make a case that WAC is a good theoretical fit within child development, neuroscience, and cognitive science literature. 

\paragraph{Neural Network Interpretation} Though not explicitly referred to was \textit{words-as-classifiers}, some recent work on neural network interpretability uses WAC-like classifiers to uncover concepts within a neural network. As is well-known, deep neural networks are `black boxes' in that it is difficult to ascertain what a neuron or set of neurons within a neural network is doing. To help uncover what neurons are doing, \citet{Kim2018-uc} introduced \textit{concept activation vectors}, which are linear (e.g., logistic) classifiers that learn to activate when certain `concepts' are observed in a network. The set of concepts is determined by a designer as important to the task. For example, if a designer wants to know if the network is picking up on features for a certain visual input, e.g., stripes, they can create a \textit{stripes} concept activation vector, which is a WAC-like linear classifier that takes in the activations of all or a subset of neurons within a particular neural network layer. Positive examples are derived from activations when the input is known to be \textit{striped}, and negative samples are derived from activations when the input is known to be something other than \textit{striped}. In this way, WAC-like concept classifiers can help uncover what neurons are learning within a network, particularly if they are contributing to a specific concept. 

\paragraph{Referring Expressions in Child Development, Fast-Mapping} Early and continued uses of WAC have focused on referring expressions to physical objects which is only a subset of language use, but it is an important one: referring expressions to physical objects is one of the most fundamental and early language uses of small children \cite{McCune2008-oy}. Importantly, the setting of learning and using referring expressions takes place in is situated, face-to-face \textit{spoken dialogue}. Caregivers and language-learning children denote objects that are visible by both dialogue participants which is the basis for learning grounded meanings (i.e., deriving a word's connotation through examples of denotation \cite{Dahlgren1976-oh}). \citet{Torres-Foncesca2022-ry} was able to model a language-learning spoken dialogue system on a robot platform that used WAC to ground words uttered by human participants to objects they referred to. WAC was able to often learn the grounded meaning of a word with a single use coupled with multiple camera frames in reference to an object (similar in a way to so-called ``fast-mapping," where children can quickly learn words with only minimal exposure; see \citet{Dollaghan1985-es}). 

\paragraph{Opposites and Contrastive Learning} Training a model using positive and negative examples has become fashionable in large-scale LM training and reinforcement learning, usually termed \textit{contrastive learning}. Learning by `opposite' inputs is examined in \citet{McGilchrist2021-yd} (Chapter 20): ``The inhibitory action of the corpus callosum enables the human condition. Delimitation is what makes something exist. Friction, for example, the very constraint on movement, is also what makes movement possible at all. In its excess, true, we are immobilised; yet so we are in its absence. There is nothing to push against. Resistance can put the brakes on motion, or cause motion; it can prevent or cause change." and he further quotes CS Pierce: ``A thing without oppositions ipso facto does not exist ... existence lies in opposition." (p.1250) While the WAC model could be trained using a different regime from positive and negative samples, the most common is a kind of contrastive learning where positive examples are contrasted with their `opposites' which, in the case of a word, simply means where the word is not used, which seems to fit theories of how concepts are learned by humans. 

\begin{figure}[ht]
  \centering
  \includegraphics[width=.4\textwidth]{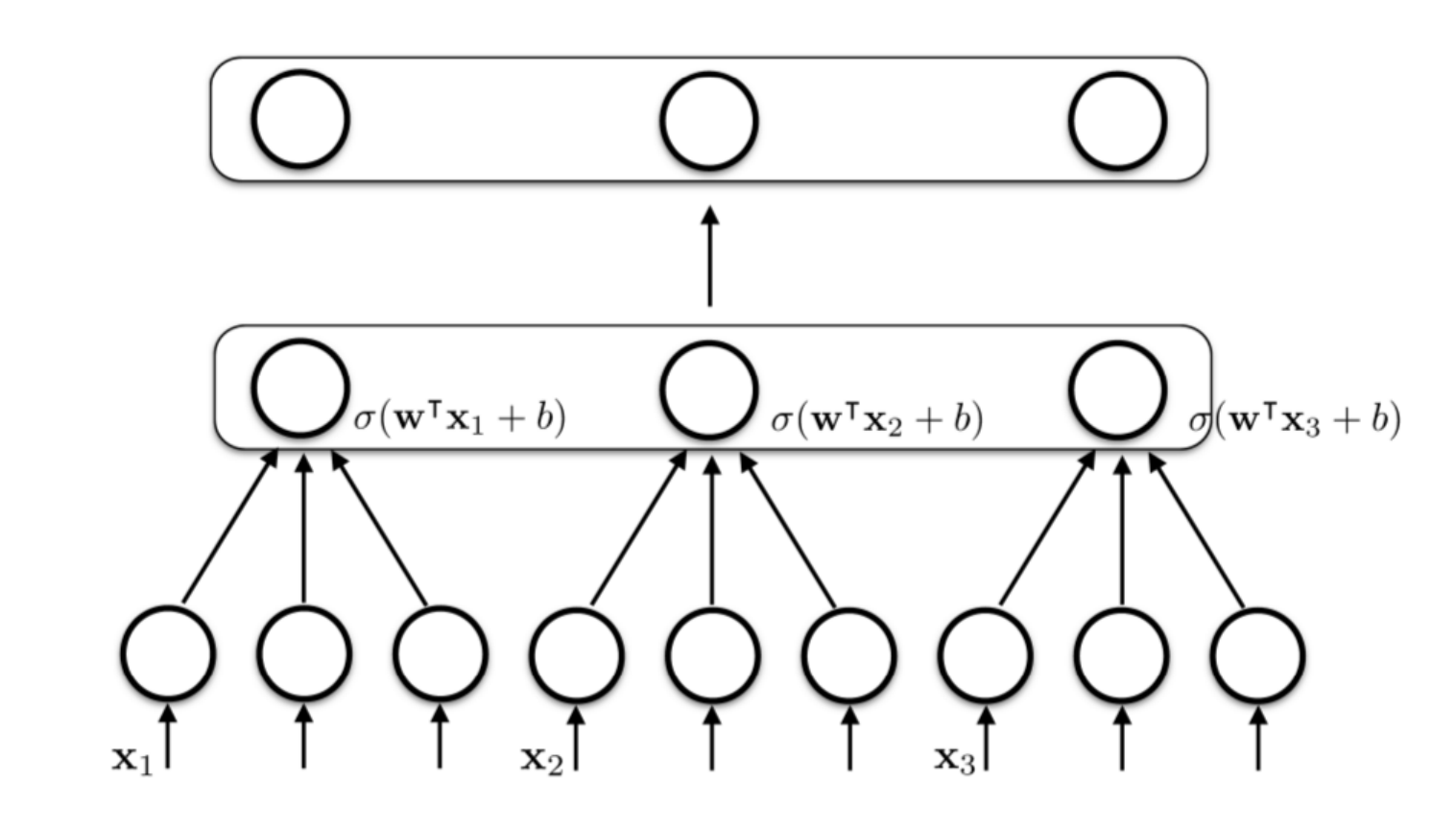}	
  \caption{Representation as network with normalisation layer, from \citet{Kennington2015-ou}.\label{fig:wac:maxent}} 
\end{figure}

\paragraph{Neurons and Concepts} The most common use of the WAC model treats words as fundamental concepts: individual word classifiers can be seen as individual neurons that learn a specific word concept. This aligns with theories of cognition that claim individual neurons activate when they are exposed to (i.e., `see') particular individuals (e.g., when one sees their grandmother, or a particular celebrity; see \citet{Quiroga2013-ru} for an overview, but also note that the single-neuron per concept theory has its critics \cite{Thomas2017-kk}). We are not claiming that neurons within human brains function in this way, but it is not controversial to say that the human brain has neurons that activate when certain input values are detected. While WAC models words as individual neurons, a WAC model could theoretically exist where word meanings are distributed across neurons (see \citet{Kennington2022-xs} for a preliminary exploration) or as part of a larger network, as portrayed in Figure~\ref{fig:wac:maxent}.

\paragraph{Cognitive Networks} WAC has been used in more complex networks in cognitive science literature. WAC was used, for example, to extend a \textit{multiplex} network \cite{Stella2017-dj} of concept categories (to ``to cognize is to categorize" \cite{Harnad2017-jo}) from different modalities (in their case, word associations, feature norms, phonological similarities, and co-occurrence features) to include cognitive layers from vision and proprioception, resulting in improved network growth that aligned with estimates of when words are learned by children \cite{Ciaglia2023-rq}. Though an early example, this has implications to larger questions about cognitive networks explored in \citet{Stella2024-fi} including how WAC and cognitive networks can enrich LMs with multimodal knowledge, some early results reported in \citet{Fields2023-yi}. 

\section{Towards Unifying Formal, Distributional, and Grounded Semantics with WAC}

In this section, we showcase how WAC, a model of grounded word-level semantics, has already been unified with formal, symbolic models and with neural, distributional models. 

\subsection{WAC in Formal Semantics}

Early versions of WAC formulations were designed to be integrated with symbolic, semantic formalisms. \citet{Kennington2015-ig} originally formulated WAC this way: 

\begin{center}
\begin{equation}
  \denotes{w}_{obj} = \lambda \mathbf{x}. p_w(\mathbf{x})
\label{eq:wac:intensobj}
\end{equation}
\end{center}

Where $\denotes{w}$ denotes the meaning of $w$, and $\mathbf{x}$ is the type of feature given by $f_{obj}$, which is something denoted like an object, the function computing a feature representation for a given object (for example, common visual features or output of a layer from a deep visual network like a convolutional neural network, as done in \citet{Schlangen2016-mz} and \citet{Torres-Foncesca2022-ry}). The features $\mathbf{x}$ are applied to the predicate $p_w$ which returns a probability, which is the \textit{fit}, or degree of believe that $p_w$ is a good fit to the object.

Though a simple formulation, this effectively links WAC to a formal representation because the functional application of an object's features to the classifier is akin to functionally applying a variable to a first-order predicate (e.g., $red(x)$, where $x$ is an object's feature representation), but instead of returning only true or false within a rigid formalism, the predicate returns a probability. This is a powerful extension of formal representations because often the formalisms are concerned with sentence-level meaning, but individual predicates (such as $red$) don't actually know when to return true or false; a classifier can be trained to return a value between and including 1 and 0, thereby giving formal semantics a degree of connection to the physical world. 

\citet{Larsson2013-po} went beyond simple, single predicates and considered a larger formal machinery that links a grounded WAC approach to \textit{Type Theory with Records} (TTR) \cite{Cooper2005-xp}. TTR is a formal representation framework for semantics, using records to express types and has been well-explored as a comprehensive framework for interactive dialogue, including functionality for handling common dialogue artifacts such as false-starts, repetitions, and clarification requests. A \textit{record} is a frame-like structure with attributes termed \textit{labels}, and a record type is a frame with corresponding attributes and values that mark the types (e.g., a logical predicate such as $red$). WAC classifiers can effectively operate in this space of replacing predicates with classifiers that return probabilities upon which TTR can operate. 

\begin{figure}
    \centering
    \includegraphics[width=0.8\linewidth]{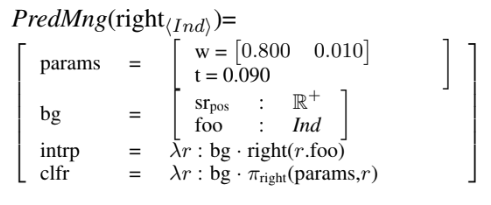}
    \caption{Example of Type Theory With Records applied to the \textit{right} classifier, from \citet{Larsson2020-gl}.}
    \label{fig:ttr}
\end{figure}

\citet{Larsson2020-gl} extended TTR with classifiers in an example of a dialogue interaction that applied the WAC probability in a real-world scenario of language understanding. A TTR example is shown in Figure~\ref{fig:ttr}, where the $w$eights (i.e., coefficients) of a classifier are part of the record for $params$ and the classifier $clfr$ for the word \textit{right} is a predicate that uses the classifier's parameters (not unlike the formulation in Equation~\ref{eq:wac:intensobj}, though more couched within a principled formal system). \citet{Cooper2023-ai} extended this line of inquiry into more detail, effectively bridging perception with formal semantics; importantly, while keeping the domain of language use within interactive dialogue. 

Though only a few examples, the work reviewed here illustrates that WAC can be integrated into semantic formalisms with provisions for handling dialogue, which can help WAC overcome its shortcomings of composition, and WAC can give the formalisms learned access to the physical world, effectively grounding formal semantics. 

\subsection{WAC in Distributional Semantics}

\begin{figure}
    \centering
    \includegraphics[width=0.6\linewidth]{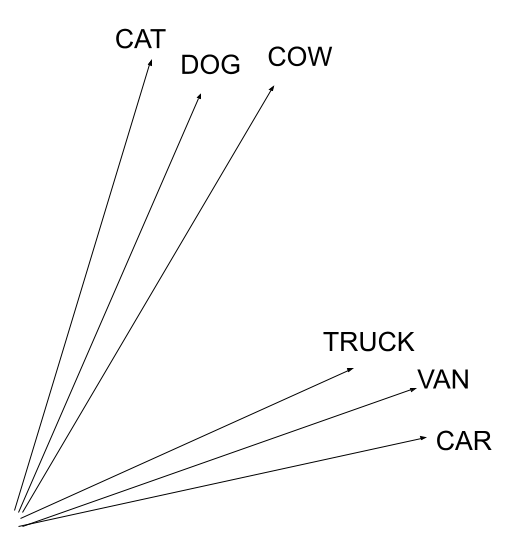}
    \caption{Example of word vectors: animal words with similar meanings and vehicle words with similar meanings cluster together, but the two groups are far from each other.}
    \label{fig:distexample}
\end{figure}

Before LMs became mainstream, \citet{Zarriess2017-sg} showed that WAC could be bridged with distributional theories using vectorized continuous representations of visual information as well as textual information. Their model also brought reference resolution to real objects within a \textit{situational context} that captured similarities between words occurring for different objects within the same visual scene. 

Mapping from a visual scene to a useable representation deserves more attention. One of the powerful abilities that deep neural networks have is that they are able to take noisy inputs and abstract over features that are meaningful to high-order concepts, such as words. Convolutional neural networks, for example, can be given raw image pixels as input and produce a distribution over abstracted object types (e.g., car, kite, bird) and it has been shown that layers within the network pick up on progressively more abstract visual concepts beginning with straight lines in a certain direction, moving to combinations of lines such as corners, until specific object attributes such as ears and tails can be used as features to determine if an object is a specific animal such as a cat or a dog. Surely, that means that neural networks are grounded?

It depends on their architecture and how they are trained. In contrast to convolutional networks, LLMs are trained largely on text, which (as explained in the Introduction; see Figure~\ref{fig:distexample} as an example of how word-level vectors are represented) is still a form of language that is not grounded. Visual LMs have become prevalent in recent years due to improved datasets and modeling approaches, but the question as to their groundedness within a semantic framework remains an open one. Most visual LMs fuse the text and visual modalities at the Attention stage \cite{Fields2023-zv}, which means that the word-level word2vec-like distributional representations (i.e., the LLM's embedding layer) are learned distinctly from any image input. For example, the LM might have an embedding for the word \textit{red} as learned from textual data, but the embedding has no notion of what redness is as perceived. In other words, the embedding doesn't know red when it sees it and is therefore ungrounded. 

\citet{Kennington2021-sj} showed that the embedding layer can be enriched with visual information. In that work, the author collected images representing the vocabulary of a LM (in their case, ELECTRA \cite{Clark2020-vk}, a vocabulary of over 30k words) and trained a WAC model, resulting in a logistic regression classifier for each word---classifiers that were trained on visual knowledge. They then extracted the coefficients out of each classifier (representing a trained, visual embedding) and replaced the text embeddings in ELECTRA with the visual embeddings from the classifier coefficients. They then pre-trained the model on text data (after freezing the visual embedding layer weights for the first part of the training regime) and found in their evaluation that adding visual information in the embedding layer helped with visual (specifically, a visual dialogue task where dialogue participants had to answer questions about visual objects \cite{Das2019-sh}) and non-visual (i.e., text only) downstream tasks. We extend this model to not just replace textual embeddings with visual ones, but combine both embeddings in a preliminary experiment which we explain below.

\subsubsection{Preliminary Experiment: WAC in Language Model Embeddings}

In this section, we explain a simple, preliminary experiment that illustrates how WAC can be applied in distributionally-trained models such as LMs. The model described in \citet{Kennington2021-sj} is an important step in showing how WAC can be unified with distributionally-trained models, but the limitation in that approach is that the visual embeddings \textit{replaced} the textual ones. In our experiment, we make use of both types of embeddings and test different methodologies for unification. 

\paragraph{Task \& Procedure}

\begin{figure*}
    \centering
    \includegraphics[width=1.0\linewidth]{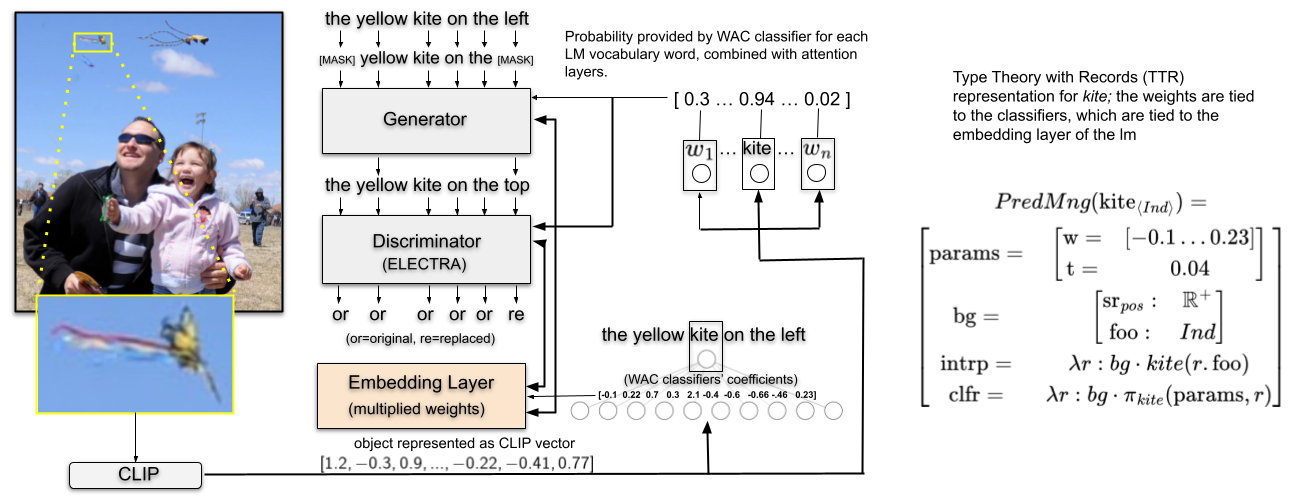}
    \caption{Depiction of extracting visual vectors and applying them to WAC classifiers, which are in turn used in two ways (1) classifier coefficients are combined in the embedding layer and as part of the TTR representation, and (2) classifier probabilities applied to objects are combined in the Attention layers. Adapted from \citet{Kennington2021-sj}.}
    \label{fig:trainingvectors}
\end{figure*}

We use the ELECTRA-small model and pre-train on a subset of the openwebtext dataset (we leave larger-scale training data and longer training regimes for future work). We train using randomly initialized weights, but use the standard training parameters for ELECTRA-small. 

In our experiment, we combine the visual embeddings used in \citet{Kennington2021-sj}, which had vectors of size of 128; i.e., there is a visual vector provided by WAC for each word and a corresponding text vector provided by the LM's embedding layer for each word. We compare three different ways of combining the visual vector for a word $\vec{v}_{v/w_i}$ with the textual vector for for the same word $\vec{v}_{t/w_i}$:
\begin{itemize}
    \item $\vec{v}_{v/w_i} + \vec{v}_{t/w_i}$ (vector addition)
    \item $concat(\vec{v}_{v/w_i},\vec{v}_{t/w_i})$ (vector concatenation---this required small changes to model hyperparameters)
    \item $\vec{v}_{v/w_i} * \vec{v}_{t/w_i}$ (vector multiplication)
\end{itemize}

The $\vec{v}_{v/w_i}$ for a word $w_i$ is a trained binary logistic regression classifier using 100 positive examples of images for $w_i$ and a randomly sampled 300 negative images from any other word. The images are transformed from raw pixels to a vector using the CLIP model \cite{Radford2021-pk}, each vector is input to the classifier for $w_i$.\footnote{Applying vectors from models that are more theoretically aligned with language learning is left for future work.} After training each classifier, we extract the coefficients and use those as the visual vector $\vec{v}_{v/w_i}$. We only train the visual vectors once and use them throughout the pretraining regime; the textual vectors change as the model is trained. This process is depicted in Figure~\ref{fig:trainingvectors} where the CLIP output is used as input into WAC classifiers; once they are trained, they are combined with the embedding layer.

\paragraph{Metrics} We use two benchmarks for this preliminary experiment: MRPC and WNLI. MRPC is a semantic similarity score that is reported as an f1 value, and WNLI is a natural language inference benchmark reported as a percentage. For a baseline, we compare pre-training ELECTRA-small using the same training data without combining a visual embedding. 

\paragraph{Results}

\begin{table}
    \centering
    \begin{tabular}{c|c|c}
        \textbf{Model} & \textbf{MRPC} (f1) & \textbf{WNLI} (acc) \\
        \hline
        baseline & 0.78  & 0.49\\
        concat & 0.78 & 0.43 \\
        add & 0.68 & 0.45 \\
        mult & \textbf{0.81} & \textbf{0.56} \\
    \end{tabular}
    \caption{Experimental results for MRPC and WNLI: multiplication produced significantly higher results than other methods and the baseline.}
    \label{tab:results}
\end{table}

Table~\ref{tab:results} shows the results of our preliminary experiment. We can see that there are no improvements over the baseline when embeddings are concatenated or added; in fact, adding vectors together results in worse results (as might be expected for adding, due to negatively changing the optimization space). However, there are noticeable improvements above the baseline when vectors are simply multiplied. Even in these text-only benchmarks, improvements are seen when fusing visual information into the word-level `semantic' (i.e., embedding) layer of a LM, showing that visual information can enrich language models, but in this case the text embeddings are not overridden---they both can learn together. We leave for future work training the visual and the text embeddings together instead of holding the visual embeddings constant throughout the training process. We conjecture that multiplication works better because both modalities are weighting each other resulting in smaller, yet meaningful changes in the optimization surface during learning.

\section{Unified Semantics with WAC} \label{sec:unify}

\citet{Schlangen2020-hv} outlined a model that used WAC, coupled with formal and word-level distributional theories where the linguistic phenomena in question take place in a dialogue setting. This overcame some of the limitations of WAC only being trained bottom-up with referring expressions; in Schlangen's model, word meaning is conceptualized as a \textit{concept file} that can contain definitional, taxonomical (e.g., a \textit{dog} is a \textit{mammal}) perceptual (through a WAC classifier), and distributional information about a word, and a learner can have definitional information about an object before they perceive it (e.g., a \textit{zebra} is like a horse with stripes; though no perceptual examples are available the concept is understood enough to discuss or recognize a zebra). This work is also couched within a formal system (the author mentions several, including TTR), effectively bridging WAC with distributional and formal theories. In this section, we extend this theory to bring WAC into the modeling framework of LMs.

The limitation with the above experiment is, while the WAC classifiers were used to produce the visual embeddings, they were thrown away, leaving the rest of the work in training and evaluation to be done by the LM. Moreover, most visual LMs fuse knowledge of the visual world at the Attention stage which has been shown to work on visual-language tasks. Is there a more direct way to incorporate WAC into LMs that both (1) brings visual semantic (connotative) knowledge into the embedding layer and (2) incorporates visual (denotative) knowledge into the Attention layer without throwing away WAC? 

We have already seen that we can use WAC coefficients as embeddings, but what about the Attention layer? This is where the actual WAC classifiers can contribute: given visual input of an object from an image, pass the object representation into each trained WAC classifier up to the model's entire vocabulary (i.e., ones that have been trained to that point). Each WAC classifier then produces a fitness probability. For example, if the image has a picture of a kite, then the \textit{kite} classifier should return a high probability, but the \textit{table} and other classifiers should return a low probability. More formally, each word classifier produces a probability $p_{w_i}$ that we can combine into a vector $[p_{w_1}, p_{w_2}, ... p_{w_n}]$ of size $n$ (the length of the vocabulary) which we combine at the (cross-)Attention level (usually multiplication with the Values matrix). 

During training, the WAC classifiers are trained in conjunction with the LM. For example, using a dataset of referring expressions made to objects in images, the referring expressions are used to train the LM (including text embeddings; similar to  \citet{Zarriess2017-sg}), and the individual words are represented separately by WAC classifiers; positive training examples from objects referred to in the dataset / negative training examples randomly selected from objects not referred to. WAC is retrained at intervals (e.g., one time every batch or dialogue interaction) to produce the new visual embeddings, and the newly trained classifiers are used in deriving the WAC probabilities for the Attention layers during training. 

Such a formulation makes WAC an integral part of the LM training, with representations in the embedding and Attention layers, with the ability to both understand the semantics of visual words as well as apply the visual knowledge to real-world visual tasks (i.e., connotative and denotative capabilities). WAC classifiers are not thrown away, but used throughout training and application; WAC can still operate on small amounts of data and be accessible to formal representations like TTR outside of the LM.\footnote{There are some similarities to our sketch and the model proposed in \citet{Mao2019-rh}, which claims a degree of grounded, neuro-symbolic capabilities for a LM (learned interactively), though like others, the embedding layer is not grounded, and the symbolic side is a simple ontology of objects and actions.} Figure~\ref{fig:trainingvectors} shows a depiction of this sketch: as in the experiment, the classifier weights are combined (using multiplication) to the textual embeddings and in TTR; each vocabulary word in the lm has a WAC classifier, the probabilities of each are combined as a denotation vector that is combined in the LM's attention layers.

Furthermore, the WAC classifiers can learn in an interactive, spoken dialogue setting with small amounts of data, as has been done in the past \cite{Hough2015-id,McNeill2020-zd,Hough2020-qf,Torres-Foncesca2022-ry,Hough2024-iu}. The setting of spoken dialogue is often forgotten in formal semantics and in data used for training LMs (TTR being a long-standing, notable exception, among others), but since the basic and fundamental site of language learning for humans is spoken dialogue, using interactive dialogue data for training can likely improve models with less data, following work in \textit{curriculum learning} where samples of simpler language use are used before more complex ones, resulting in LMs learning effectively with less data \cite{Xu2020-we}.\footnote{Though trends towards using reinforcement learning in interactive settings such as GRPO \cite{Shao2024-eh} and RLHF \cite{Ouyang2022-ho} are steps in the right direction, they are often applied after non-interactive pre-training.}


\section{Discussion \& Conclusion}

WAC is not a complex model because it operates at the word level, but modeling word-level semantics is necessary for all semantic theories. WAC is a straight-forward grounded model that has a track record of being used in many domains, grounding with many different modalities, being applied in spoken dialogue settings, learning from small amounts of data, and effectively being unified with different semantic theories. Even though models of language have become more complex, since WAC was introduced 10 years ago, the theoretical properties and practical uses have shown that WAC has made an important contribution over the years. 

We see from the literature reviewed above, and preliminary experiment that (1) WAC can be used as a way to unify grounded and formal semantics, and (2) WAC can be used to unify grounded and distributional semantics. Moreover, being a classifier with coefficients, WAC can operate in both probability and linear algebra space, making WAC a model that can easily integrate with deep neural networks including LMs. Given this review, our small experiment, and our unified model sketch, in our estimation, the road to grounded, neuro-symbolic AI is indeed paved with words-as-classifiers. 

In our current and future work, we are attempting to implement and evaluate WAC in an LM architecture like the one explained in Section~\ref{sec:unify}.

\bibliography{paperpile}
\bibliographystyle{acl_natbib}




\end{document}